# INTEL-TUT Dataset for Camera Invariant Color Constancy Research

Caglar Aytekin, Jarno Nikkanen and Moncef Gabbouj, Fellow IEEE

*Abstract*—In this paper, we provide a novel dataset designed for camera invariant color constancy research. Camera invariance corresponds to the robustness of an algorithm's performance when run on images of the same scene taken by different cameras. Accordingly, images in the database correspond to several lab and field scenes each of which are captured by three different cameras with minimal registration errors. The lab scenes are also captured under five different illuminations. The spectral responses of cameras and the spectral power distributions of the lab light sources are also provided, as they may prove beneficial for training algorithms to achieve color constancy. For a fair evaluation of future methods, we provide guidelines for supervised methods with indicated training, validation and testing partitions. Accordingly, we evaluate a recently proposed convolutional neural network based color constancy algorithm as a baseline for future research. As a side contribution, this dataset also includes images taken by a mobile camera with color shading corrected and uncorrected results. This allows research on the effect of color shading as well.

## I. INTRODUCTION

COLOR constancy is a feature of the human visual system (HVS) that allows perceiving the color of the same object relatively constant under different illuminations [1]. This trait is remarkable given that the color constancy problem is in fact ill-posed. The ill-posedness follows from the fact that both object reflectance properties and light source illumination properties are unknown, yet the only available information is the perceived color. It is argued that the HVS achieves color constancy by first approximating the composition of the illuminating light and then discounting this estimated illumination in order to obtain the true object color [2]. Many computational color constancy (CC) methods also follow this approach in either a supervised or unsupervised manner.

Unsupervised methods can be grouped into two categories based on the different strategies used while estimating the color of the illuminating source. First category makes assumptions on the reflectance statistics in a scene. These methods commonly assume a Lambertian surface exhibiting an ideal diffuse reflection, i.e. light is reflected equally among all the directions. In particular, White-Patch [3] assumes that there is a perfectly reflecting material in the scene, whereas Gray-World [4] assumes that the average reflectance in a scene is gray, i.e. achromatic. Shades of Gray [5] assumes that the average reflectance in a scene is gray when raised to the power of $p$ and empirically finds the best performing $p$ value. It has been further shown that White-Patch and Gray-World methods can be represented in the same formulation with $p = \infty$ and $p = 1$, respectively. Rather than the average reflectance, Gray-Edge [6] method assumes that the average reflectance differences in a scene are achromatic. This work also proposes a new general formulation and shows that the methods in [3] - [6] are instantiations of this formulation.

The second category of unsupervised methods makes use of the physical property of objects in the scene, and can thus be called physics-based methods. These methods are not restricted to the Lambertian surface assumption of objects, but they also make use of specular reflection: a mirror-like reflection of the light where the reflected light is in the direction, which makes the same angle to the surface normal as that of incoming light. It was observed in [8] that from the color histogram of a surface with a constant diffuse color, one can obtain diffuse and specular components of the reflected light. The seminal work in [7] shows that specularity can be used to compute illumination. When the reflected light from a surface involves both diffuse and specular components, it was observed that in CIE chromaticity diagram, the points corresponding to the same surface fall on a straight line that passes through the specular point. The results of this work were later used to robustly estimate the illuminant chromaticity via the information from the specular components [9],[10].

The supervised approaches to CC can also be categorized into two types. The first category tries to learn a combination of the unsupervised methods to estimate the illumination. In [11], this was achieved by first obtaining several estimates of the illuminant via unsupervised methods, and then selecting the one that results in the most likely semantic content of the image. In [12], images were classified as indoor or outdoor and then separate models were used for illuminant estimation, which are learnt for each category by tuning and combining the statistical

C. Aytekin is with the Department of Signal Processing, Tampere University of Technology, Korkeakoulunkatu 1, 33720, Tampere, Finland (e-mail: caglar.aytekin@tut.fi)

J. Nikkanen is with INTEL, Insinöörinkatu 41, 33720 Tampere, Finland (e-mail: jarno.nikkanen@intel.com)

M. Gabbouj is with the Department of Signal Processing, Tampere University of Technology, Korkeakoulunkatu 1, 33720, Tampere, Finland (e-mail: moncef.gabbouj@tut.fi)

methods. A decision forest was learned in [13] which predicts the proper CC method to be applied given a set of color and texture features of an image. The authors also propose a way to combine several methods according to the output of the classification obtained from the decision forest.

The second category in supervised techniques build their own model to learn the illumination directly. One of the earliest endeavors in this category [14] makes use of the observation that under a standard illuminant, the colors that are observed in a scene are limited and called the canonical gamut. These colors can be learned via a training procedure. Then one can apply a transformation on a test image that will map the image gamut to the canonical gamut. The illuminant can then be estimated via the resulting mapping. An extension of this method [15] shows that gamut mapping performs better when using image derivatives instead of image intensities. Popular machine learning methods have also been applied to CC based on neural networks [16], support vector machines [17], Bayesian framework [18], [19] and learning from exemplars [20]. Recently, deep learning based approaches have also been applied and shown considerable success in CC [21]-[23]. More detailed surveys on computational CC algorithms can be found in [24],[25].

Most of CC techniques aim to first solve for the projection of a canonical illuminant to the sensor spectral sensitivities and then correct the image color accordingly. For example, when the Lambertian diffuse assumption on the surface reflectance of a given scene is used, the image value $\rho_k(x)$ on a channel $k$ and a given location $x$ can be formulated as follows.

$$\rho_k(x) = \int E(\lambda)S(\lambda, x)R_k(\lambda)d\lambda, \qquad (1)$$

where $E$ is the illuminant spectral power distribution, $S(\lambda, x)$ is the surface reflectance of the object at location $x$ at wavelength $\lambda$ and $R_k(\lambda)$ is the $k^{th}$ sensor's sensitivity at wavelength $\lambda$. Given $\rho_k(x)$, CC methods aim to estimate the chromaticity of the illuminant.

$$\rho_k^E = \int E(\lambda)R_k(\lambda)d\lambda. \qquad (2)$$

One can observe from equations (1) and (2) that both the input and the output depend on sensor sensitivities, and their relationship is non-linear and depends on the surface reflectance.

CC methods do not atone for the sensor sensitivities, which may raise questions about their robustness to changing sensors. Supporting this claim, it has been shown in [26] that the performances of unsupervised methods on images of the same scene show a variation depending on the camera.

This effect is even more evident in the direct supervised methods as they would inherently model the sensor sensitivities during training. One can thus expect large variations in these algorithms' performances when the training, validation and test sets contain images taken by different cameras. This was in fact experimentally shown via evaluating the performance of a learning-based method in an inter-camera experiment [27].

All available CC datasets contain images taken by a single camera, except for [26]. Hence, the common approach to evaluate supervised methods follows training, validating and testing in the same dataset. In this case, inter-camera performance is simply not evaluated. This is an important problem in CC and we refer to it as camera-invariance.

In this paper, we provide a new Intel-TUT image database[1] for testing camera-invariance in CC. To this end, we have collected images from three different cameras that capture scenes of printouts and real scenes both in lab and field images. The lab scenes also contain several illumination conditions. The proposed database consists of 1536 images, augmented with a 2nd field test set containing 454 images taken by one of the three cameras only. We also share the spectral responses of the camera sensors as well which we find useful for future research to achieve camera invariance.

The rest of this paper is organized as follows. First, we review the available CC datasets in Section II and highlight the contributions of the proposed dataset. Next, we describe the dataset and how it was created in Section III. We also propose several recommendations to follow for evaluating the performance of CC algorithms using this dataset as a benchmark in this section. In Section IV, we evaluate the performance of the baseline CC models and discuss the results. We also propose and evaluate a convolutional neural networks based color constancy method as a baseline for future research. Finally, Section IV.B concludes the paper.

## II. COLOR CONSTANCY DATASETS

One of the earliest datasets for color constancy (CC) was provided in [28]. To capture the images in this dataset, a Sony DXC-930 3CCD camera was used. The dataset includes the spectral response of the camera, provided in steps of 4 nm within the interval 380-780 nm. In addition, a total of 1995 surface reflectance coefficients covering a variety of surfaces, and illuminant spectra, with 11 lab illuminants and 81 real-world illuminations, are also included in the dataset. The dataset also includes additional images capturing 50 scenes, with minimal specularities, non-negligible dielectric specularities, and metallic specularities, along with fluorescent scenes. Each scene was captured under 11 lab illuminants and after removing images that include calibration deficiencies, 529 images were provided. For CC techniques that assume diffuse reflectance, the usable part of this dataset includes only 30 scenes out of the 50 provided. Although this dataset shares illuminant spectra from real-world illuminations, the captured images do not include outdoor images and are restricted to the ones lit under lab illuminants.

The hyperspectral images provided in [28] contain only lab images. In [29], hyperspectral images of real-world images were collected by a progressive-scanning monochrome digital camera Pulnix TM-1010 within the range of 400-720 nm sampled at 10 nm intervals. 30 scenes were captured from rural and urban areas. Together with the images of the scenes, light reference images were also captured by the same apparatus but pointed at a uniformly reflecting flat surface. Noise and transmission effects were later corrected and finally, spectral reflectance of each pixel was extracted via normalizing the corrected image with a white standard.

A large dataset was provided in [30] that contains 11000 images of indoor/outdoor scenes shot by a Sony VX-2000 digital video camera. The scenes were captured in two locations to include variations in geography and weather conditions. The scene illumination was measured by a smooth gray sphere. The shortcoming of this dataset is that it includes low-resolution images that were subject to correction. Moreover, it was claimed in [19] that since the images were frames extracted from video sequences, only ~600 of them are truly uncorrelated.

A wide variety of images was collected in the dataset provided by [19], which contains 246 indoor and 322 outdoor images. The scenes were captured by Canon 1D and Canon 5D DSLR cameras. Although this dataset includes 2 different cameras, the same scenes were not captured by both cameras, moreover, only 86 images were taken by Canon 1D. The images contain clipped pixels, are nonlinear and demosaiced. To overcome these problems, a reprocessed version of this dataset is provided in [31], which includes almost RAW linear images with the least amount of processing.

Another dataset with 105 scenes was captured in [32] with a Nikon D700 camera. Up to 9 images of different exposures (with 1 exposure value difference between consecutive images) were captured using the camera's auto-bracketing. The raw images are then processed in two different ways. First, almost-raw base images were created, one image per exposure. Next, a set of high dynamic range (HDR) images were created from the base images. To measure the illumination ground-truth, 4 color-checkers were placed in the scene with varying angles and the median of measurements were used. The images with a high-level variation in color-checker measurements were discarded as the dataset targeted uniformly illuminated scenes.

In contrast, the dataset in [33] was specifically designed to contain multiple illumination in a scene. The dataset contains 9 outdoor and 59 lab images taken by a Sigma SD10 camera with Foveon X3 sensor. The ground-truths for outdoor images were collected with the several gray balls. Indoor images were constructed by illuminating 7 different objects using a combination of two different halogen lamps under 4 color filters. Misaligned images of the same scenes were discarded. Ground truth for lab images were collected by gray cards.

Another dataset for multiple illumination was provided in [34] containing 20 real-world and 58 lab images taken by a Sigma SD10 camera with Foveon X3 sensor. Lab images were constructed by 6 different pairs of illuminations of 10 scenes. Again, misaligned images were removed. The real-world images were carefully selected to capture scenes with 1 direct and 1 ambient light. Ground-truths were collected by a complicated procedure explained in [34].

A dataset for exploiting videos for CC was provided in [35]. The dataset includes both single-illuminant and double-illuminant videos. A Panasonic HD-TM 700 was used to capture videos. Single illuminant dataset includes 3 outdoor and 6 indoor scenes. The videos were recorded by moving the camera and the ground truth is collected via a gray card present in each video by averaging the measurements within the video. In the double-illuminant dataset, the scenes were lit by two sources and two gray cards were present in the scene where each gray card was lit by a single illuminant.

All of the above datasets are collected by a single camera with the exception of [19]. However, in the latter, the cameras, namely Canon 1D and Canon 5D are quite similar and do not capture the same scene. Besides, there is a large disproportion in the images taken by Canon 1D -86 images- and Canon 5D - 482 images. Hence, none of the above datasets is suitable for camera-invariant CC research. For supervised approaches, one might consider training and testing on different datasets for camera invariance. However, in this setup, the impact from scene and illumination variation cannot be distinguished from the impact of camera response variation. Therefore, it would not be possible to know if the camera invariance problem was in fact addressed at all. A database consisting of aligned images of the same scene taken by various cameras is needed to allow testing different hypotheses about achieving camera invariance. The only dataset that provides such images is the NUS dataset presented in [26]. The dataset includes over 1600 indoor and outdoor images taken by 9 different cameras. Each scene was captured by each camera with slight misalignments. Ground truth was measured via color-checker for each image.

The proposed Intel-TUT database is similar to NUS dataset in the sense that it also provides images of the same scene that are captured by different cameras. Moreover, Intel-TUT database has the following novelties:

- The spectral sensitivities of the cameras are provided;
- The spectral power distributions of the light sources used in lab images are provided;
- In the lab images, the same scene was captured under different illuminants.
- Unlike other datasets, a mobile camera is used in our camera set. This adds value to the dataset since mobile cameras are nowadays the most frequently used type of cameras.
- Images taken by mobile camera come with and without color shading correction, enabling further studies on the impact of residual color shading.
- We provide a field test-set which is collected by one of the cameras for a better assessment of CC methods.
- We share per-image color conversion matrices, which can be used as ground truth for light source spectrum estimation.
- Finally, in our database, no color-checkers (unless intentionally placed), gray balls or tripod legs (which could serve as a ground truth clue) are visible.

The details of the proposed dataset and the motivation for the above contributions are further discussed in the next section.

**Table 1.** Light Source Properties

| Name | IE D65 | IE D50 | IE F11 | IE F12 | IE A | SL Daylight (D) | SL Cool White (CW) | SL Horizon (Hor) | SL A | SL TL84 |
|------|--------|--------|--------|--------|------|-----------------|--------------------|------------------|------|---------|
| Lux | 1300 | 1200 | 1700 | 1400 | 400 | 1600 | 1450 | 1400 | 1570 | 1700 |
| CCT | 5950 | 4800 | 3800 | 2850 | 2800 | 6700 | 4100 | 2330 | 2840 | 3870 |
| CIEx | 0.32 | 0.35 | 0.39 | 0.44 | 0.45 | 0.31 | 0.38 | 0.495 | 0.45 | 0.39 |
| CIEy | 0.34 | 0.35 | 0.38 | 0.40 | 0.41 | 0.32 | 0.39 | 0.417 | 0.41 | 0.39 |
| Type | Fluorescent | Fluorescent | Fluorescent | Fluorescent | Tungsten/ Halogen | Tungsten/ Halogen + filter | Fluorescent | Tungsten/ Halogen | Tungsten /Halogen | Fluorescent |

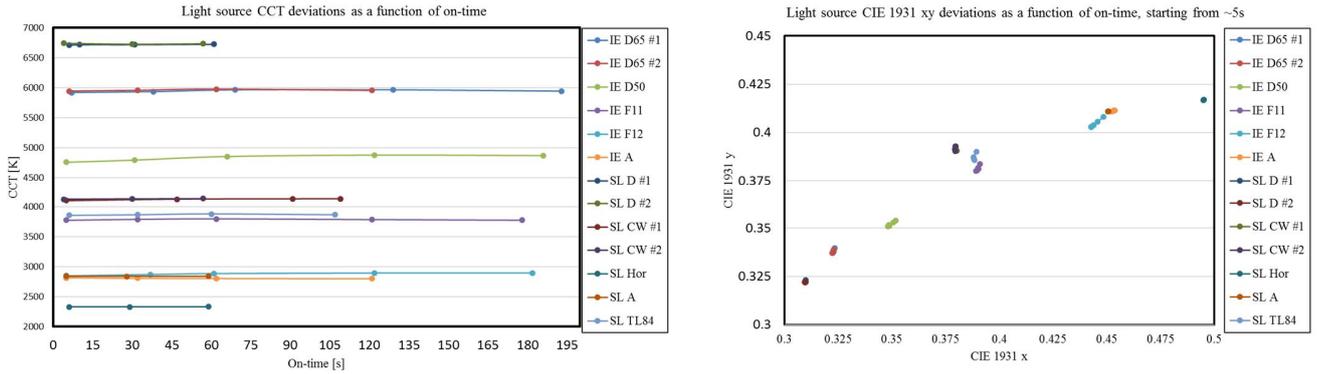

**Fig. 1** Light Source CCT and CIE xy deviations as a function of time.

## III. The Proposed Intel-TUT Dataset

The new dataset is aimed to enable research on camera invariant computational color constancy, i.e. research on algorithms that produce consistent results in the presence of changes in the camera properties. This is important especially for machine learning based methods [11]-[23]. Also, edge based methods [6] can be affected by changes in optical properties and resolution.

Machine learning based methods can be impractical if supporting a new camera module requires capturing a lot of training images. This dataset mainly aims to facilitate studies to overcome this limitation.

Images in the dataset are provided with minimal processing. This processing includes cropping the light-shielded pixels, black-level subtraction and saturation point stretching. Accompanying the raw images, we also share the corresponding JPEG files that illustrate corrected images according to the ground truth. This JPEG image has been processed with black level correction (BLC), color shading correction when necessary (CSC, only for 1 camera), white balancing (WB), color conversion from sensor RGB to sRGB (CCM), 0.45 gamma correction, minor sharpening and brightness normalization.

### A. Light Sources

The light sources used in the lab include 5 Image Engineering LightSTUDIO (IE) and 5 x-rite Macbeth SpectraLight (SL) sources. The luminous emittance (Lux), correlated color temperature (CCT), the CIE xy chromaticity and the type of the sources are presented in Table 1.

The light sources' CCT and CIE xy chromaticities deviate as a function of time, thus we have kept the light source on for at least 120 seconds before starting image capturing as we have observed that all light sources' CCT and CIExy almost stabilize after 120 seconds. Still, in order to take light source drifting into account after 120 seconds, the white points were annotated separately for different image captures under the same illumination. The deviations in CCT and chromaticities with time are illustrated in Fig. 1. Some sources were measured twice, as indicated in the figures by the measurement number next to the light source abbreviation.

The spectral power distributions of the light sources were also measured 10 minutes after switching the light source on for a more stable measurement. As it can be observed from Fig. 2, the light sources used enjoy a variety of SPDs which makes the illuminant chromaticity estimation challenging for lab images.

**Table 2.** Camera Properties

| Camera Name | **Canon EOS 5DSR** | **Nikon D180** | **Mobile Camera** |
|---|---|---|---|
| Resolution | 52 Mp (8896x5920) | 36 Mp (7380x4928) | 8 Mp (3264x2448) |
| Focal Length (35 mm eq.) | EF 24-105/4L @ 28 mm | AF-S 24-70/2.8G ED @ 28 mm | 30.4 mm (actual 4.12 mm) |
| Aperture Size | F8.0 | F8.0 | F2.4 |
| Pixel Size | 4.14 μm | 4.88 μm | 1.12 μm |
| Raw Data Bit Depth | 14 bpp | 14 bpp | 10 bpp |
| Saturation Point | 15380 | 16383 | 1023 |
| Black Level | 2046 | 601 | 64 |

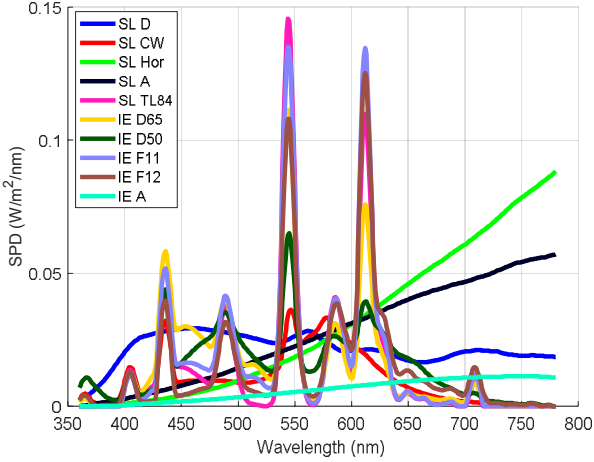

**Fig. 2** Spectral Power Distributions (SPDs) of the light sources.

### B. Cameras

Three cameras are used in Intel-TUT dataset, two high-end DSLRs, namely Canon 5DSR and Nikon D810, and one typical camera used in smartphones and tablets, which will be referred to as Mobile. The properties of the cameras are given in Table 2.

Note that the focal lengths of the DSLRs were set to 28mm, as it provided a field of view that was the closest to the mobile device field of view, and that was easy to set consistently based on the markings on the objectives. Moreover, a smaller aperture was used for DSLRs to reduce the depth-of-field difference compared to the mobile camera.

For color correction, the diagonal model of illumination change [36] is not perfectly accurate for the used cameras. That is, applying white balancing gains only corrects the achromatic colors accurately. Therefore, a vectorized 3x3 color conversion matrix (CCM) from sensor RGB to sRGB is provided separately for each image in *[filename.ccm]*. CCM characterization was made for 5 SpectraLight and 5 LightSTUDIO light sources. For the field images, the recommended CCM was annotated manually by selecting one of these 10 CCMs based on the estimated light source type. The characterization was made for each of the 10 light sources and the three cameras by minimizing the average CIE $\Delta E_{00}$ of the 18 chromatic patches of the Macbeth 24 patch chart, given a raw image captured under the specified illumination. Note that CCMs for each image is mainly provided for illustration purposes.

Color shading (CS) is the spatial variation in the chromaticity

$[R/G, G/B]$ response of the camera. It is especially evident in the raw data of mobile camera modules due to low height/high chief ray angle of the camera module. This results from the fact that light hits the peripheral parts of the IR cutoff filter and the pixel array in at a steep angle. Color shading changes as a function of illumination spectrum; especially differences in NIR energy lead to differences in color shading profile. Except in the extreme image corners, color shading of DSLR cameras was observed to be negligible. However, it was significant in the mobile camera, especially under incandescent illumination. Therefore, we provide both CS corrected and uncorrected images for the mobile camera but not for DSLR images. In Fig. 3, the red and blue channel gains for CS corrected images are illustrated. The images capture an integrating sphere to ensure a diffuse white reflectance under a tungsten/halogen light that is different from the ones in Fig. 1. The green channel is chosen as a reference since it is the central channel. As one can observe, CS effect is negligible in DSLR cameras as most of the gain deviates at most ∓1% from 1. We also observe that Nikon camera is the most robust against CS. On the other hand, mobile camera is the most susceptible as the gains deviate up to ∓5% from 1.

The CS corrected mobile images can be considered as if the images were taken by another sensor which have no CS effect. This enables research on algorithm robustness against residual color shading. Note that only the color shading is corrected to ensure the camera's $[R/G, B/G]$ response is uniform across the image. This means that lens vignetting –additional luminance falloff– still remains in the data. Correcting also lens vignetting would lead to much more clipping of bright data on the peripheral image areas.

As another contribution of Intel-TUT dataset, we provide the camera spectral sensitivities (CSS) (see Fig. 4). As one may observe, the different CSSs contain dissimilarities, which are sometimes significant. CSS may be useful for future research on camera invariance.

The light source SPDs and CSSs are validated as follows. We first calculate spectral reflectance (SR) of a white patch and use another spectrometer to measure the SPD of a light source illuminating this white patch, then take a photo of the white patch with a camera. SPD, SR and CSS are element-wise multiplied and summed afterwards per channel. The resulting chromaticities are then compared to the captured ones with the camera. The deviations are illustrated in Table 3.

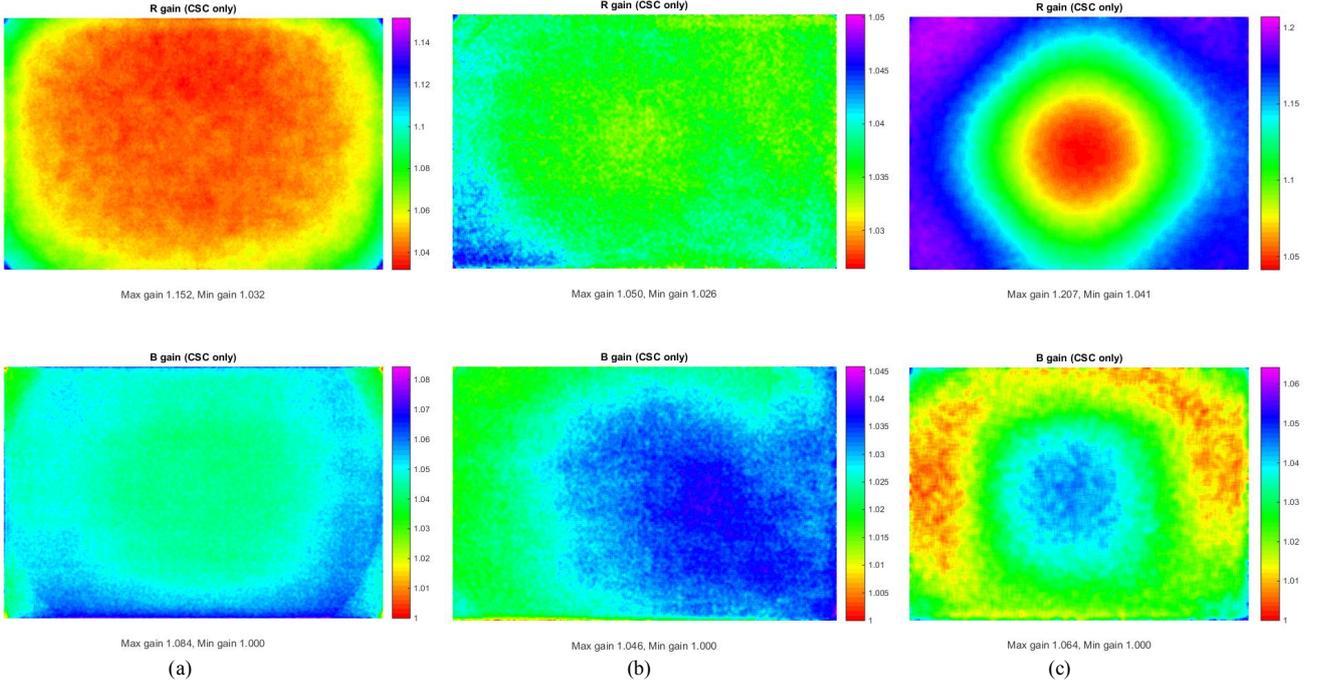

**Fig. 3** Gains of CSC corrected channels R and B for (a) Canon 5DSR, (b) Nikon D180 and (c) Mobile Camera

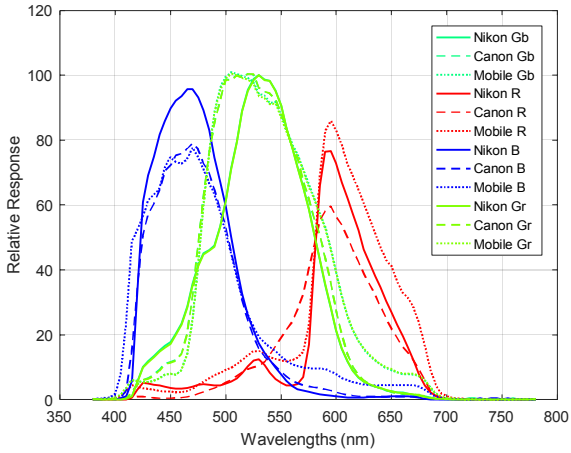

**Fig. 4** Camera Spectral Sensitivities (CSSs) of the light sources.

**Table 3**. Deviation of captured chromaticities from calculated ones.

|         | R/G    | B/G    |
|---------|--------|--------|
| SL D    | -0.7%  | -1.4%  |
| SL CW   | 1.9%   | -3.5%  |
| SL Hor  | 1.2%   | -3.3%  |
| SL A    | 0.8%   | -2.3%  |
| SL TL84 | 3.1%   | -1.2%  |

### C. Scene Contents

The scenes that are captured by the images in Intel-TUT dataset consist of lab and field scenes. Lab scenes consist of four different real scenes, in addition to one scene of 60 printouts. The field images of Intel-TUT dataset capture 64 scenes. For lab images, the illumination and the scenes are exactly the same for each camera and image framing is almost the same. For the field images, there may exist some illumination and scene content variation due to uncontrolled environment, although attention was paid to preserving nearly the same settings.

The lab real scene images represent 4 scenes, each is captured under 5 different illuminations. Only in one scene, LightStudio sources are used, whereas the others are illuminated by SpectraLight sources. The scenes are illustrated in Fig. 5.

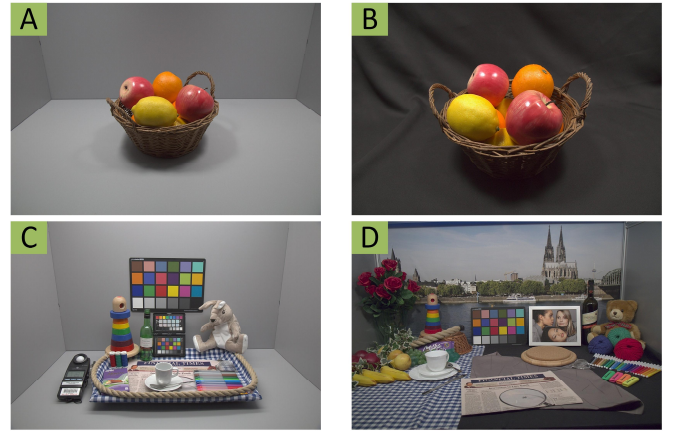

**Fig. 5** Lab Real Scenes

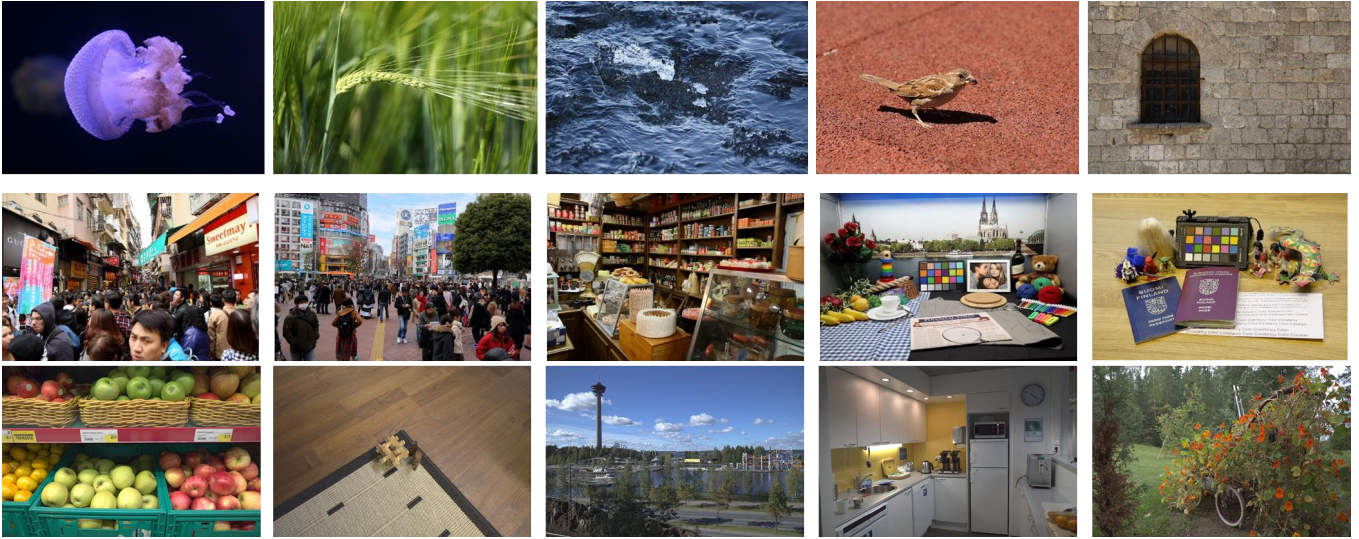

**Fig. 6** Representatives of difficult (1st row) and easy (2nd row) lab printout images and field images (3rd row).

Scenes of lab printouts aim to cover many aspects of AWB testing with a limited amount of image contents. Image printing, recapture and annotation efforts set the limit on feasible amount of images. The main purpose of printouts is not to make a simulation of real scenes, but to introduce different color distributions to images which results from different scene contents in the original scenes that the printouts were taken from. The easier scenes in this set consist of printouts representing real scenes with many different chromaticities, including achromatic objects, within the field-of-view. Difficult samples in this set include printouts representing real scenes with a limited amount of chromaticities with no achromatic objects included and ambiguous chromaticities present. Representatives from both easy and difficult images are illustrated in Fig. 6. The contents in printouts include persons, landscapes, cityscapes, animals, architecture, etc., things that are typically present in photographs. Strong mixed illumination content, color inconstant reproductions (intentional rendering with some chromatic color cast), and intentional artistic colorcasts are avoided in order to prevent ambiguous interpretations of deviations from the ground truth. In the re-captures, printouts are only illuminated with one light source and no reflection from the room is present, so the provided ground truth white point is both accurate and unambiguous. In order to minimize specularities due to the printed-paper material, the light is ensured to illuminate the printout at a steep angle. From the recaptured printouts, all image areas outside the printout has been masked from the raw Bayer images with values 0. Note that 0 is much smaller than the black-level, hence these areas can be identified easily. Masking is done with the largest rectangle that does not contain any background region with value 0.

In order to provide an evaluation of how strongly printouts represent real scenes, we illustrate a case where both the real scene and its printout are captured. The corresponding real scene is the one in Fig. 5-D and the printout of the same scene is illustrated in Fig. 6, 2nd row, 4th column.

The field images were captured with different cameras mounted on a stationary tripod. Since the field images are taken under uncontrolled conditions, small variation in illumination and scene contents may exist between different cameras' outputs. Especially in half cloudy weather conditions, the moving clouds can cause variation in the prevailing illumination within a small period of time. Moving cars and people can cause changes in the scene. Extra effort was taken in order to minimize these variations between camera captures. Representatives of field images are illustrated in Fig. 6. It can be observed that field images include a high variety of images from both indoor and outdoor scenes. There are images with a high amount of chromaticities and others with a limited number of chromaticities. Some images include more than one light source and the scenes sometimes contain specularities.

Intel-TUT dataset includes a 2nd field set containing 454 images taken from different parts of the world including Finland, Israel, Netherlands and USA, taken by Canon 5DSR.

### D. Ground Truths

The ground truth or the white point, meaning the chromaticity of the illumination in terms of camera response, is annotated separately for each image in the dataset. The white point is then stored as $[R/G, B/G]$ coordinate. The ground truths are obtained via Macbeth color-checker by averaging the readings from the achromatic patches of the checker. The gray patches in the checker are numbered as #19-#24, from bright to dark. The darkest patches are omitted for ground truth calculation and typically, patches #19-#22 are used. In some cases, #19 and #20, the brightest patches, were omitted if their readings were too close to saturation.

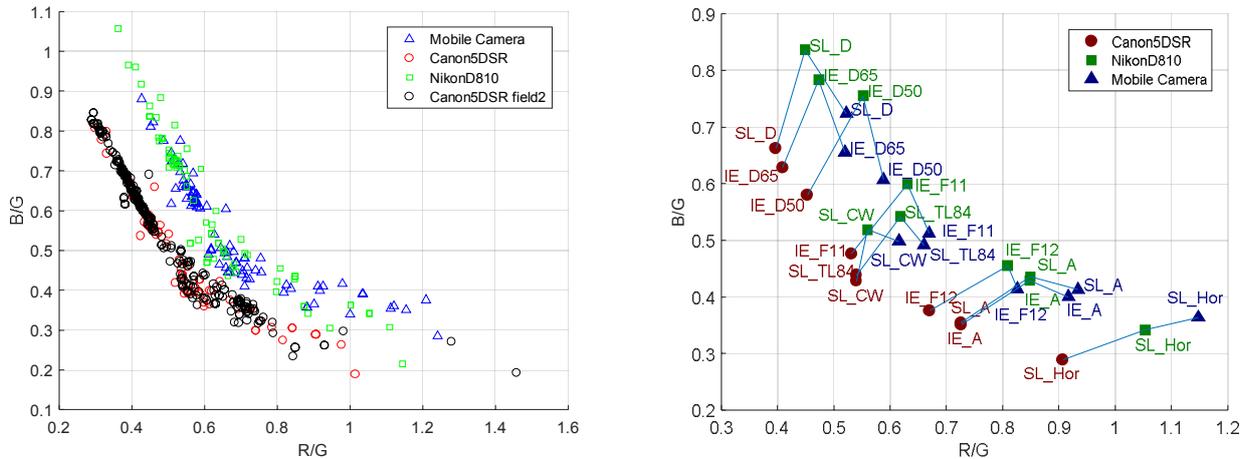

**Fig. 7** (a) Distributions of the white points in the Intel-TUT dataset, (b) camera responses to the same light source.

It should be noted that the color-checker is photographed separately from the image of the scene in order not to include the ground truth information within the image data, yet under the same illumination. Only in some images, the color-checker is intentionally included in order to populate the colors in the scene.

The distributions of the white points in the Intel-TUT dataset is shown in Fig. 7-a and it illustrates the variety of the light sources in the dataset and the effect of the camera sensitivities. Notice that there are two curves for Canon5DSR camera, one for the images from the 2nd field images and the other for the remaining images in the dataset.

Fig. 7-b illustrates the different camera responses to the same light sources. As one may observe, cameras' responses are quite dissimilar. Moreover, this dissimilarity does not follow the same pattern in different locations of the chromaticity diagram. Note especially the fluorescent light sources SL_CW, SL_TL84, and IE_F11 that have very peaky spectral power distributions. In other words, this pattern is also dependent on the illuminant, which makes the camera invariance problem even more challenging.

### E. Recommendations for Evaluating CC Algorithm Performance

In order to use Intel-TUT database as a benchmark for evaluating the performance of future CC algorithms, we recommend that the following strategies be adopted:

#### 1) Camera Invariance

We propose a six-fold cross-validation where in each fold training, validation and test sets include images taken from separate cameras. In case a validation set is not required, we suggest using two types of cameras for training and one for testing, i.e. a three-fold cross-validation. The 2nd field set can be completely ignored in this evaluation. Note that this strategy might suffer from scene memorization as the scenes in the training, validation and test sets are indeed the same. Nevertheless, it provides a fair evaluation of an algorithm's camera invariance, i.e. generalization to a new camera.

#### 2) Camera and Scene Invariance

In this case we propose using Nikon and Mobile camera images for training and Canon images for validation. For the test set, we suggest using the Canon images in the 2nd field set. This strategy helps achieving camera invariance during validation and provides an evaluation of generalization to new scenes during testing.

#### 3) Camera and Scene Invariance from Single Camera

We suggest using random partitions from the 2nd field set for training and validation and images taken by Nikon and Mobile cameras only for testing. In this way, we can evaluate if a supervised CC algorithm can learn both camera and scene invariance from images taken by a different single camera (Canon).

#### 4) Testing the Effect of Color Shading

For Recommendations 1-3, we discourage using images taken by mobile camera that have not been processed for color shading (CS) correction since these images do not correctly reflect the illuminant chromaticity consistently across the image. However, to observe the effect of CS, one can compare the results on CS-corrected and CS-uncorrected images only when mobile camera images are used in the test set.

#### 5) Testing the Effect of Resolution

Some color constancy methods may benefit from the high resolution of a digital single-lens reflex camera (also called a digital SLR or DSLR) and may produce inconsistent results in a mobile camera. In order to investigate this effect, we also share a downscaled version of the dataset including 1080p images. One may conduct an additional experiment in order to observe the changes in performance when using 1080p images. This enables a means of evaluating the algorithm's dependency on resolution; however, this comes with the cost of information loss due to bilinear downscaling.

**Table 4**. Baseline Performances on High Resolution Images: Mean, Median and Maximum Recovery Angular Errors of WP [3], GW [4], SoG [5] and GE [6].

| Camera Name | Canon 5DSR | | | | Nikon D180 | | | | Mobile (With CSC) | | | | Mobile (No CSC) | | | |
|---|---|---|---|---|---|---|---|---|---|---|---|---|---|---|---|---|
| Algorithms | WP | GW | SoG | GE | WP | GW | SoG | GE | WP | GW | SoG | GE | WP | GW | SoG | GE |
| Lab Printouts | | | | | | | | | | | | | | | | |
| Mean | 3.67 | 5.11 | 3.76 | 2.69 | 3.59 | 5.88 | 4.26 | 2.97 | 2.31 | 4.55 | 3.54 | 3.81 | 2.34 | 4.55 | 3.49 | 3.74 |
| Median | 2.72 | 4.26 | 2.92 | 2.17 | 3.29 | 4.81 | 3.25 | 2.44 | 1.90 | 3.59 | 2.87 | 2.74 | 1.88 | 3.63 | 2.71 | 2.81 |
| Max | 19.99 | 17.00 | 16.25 | 14.37 | 12.60 | 18.84 | 19.44 | 16.11 | 8.34 | 15.08 | 15.04 | 19.61 | 8.66 | 15.02 | 14.72 | 18.39 |
| Lab Real Scenes | | | | | | | | | | | | | | | | |
| Mean | 6.94 | 3.96 | 4.71 | 4.93 | 9.50 | 4.91 | 5.14 | 4.39 | 7.62 | 3.42 | 6.62 | 6.32 | 8.67 | 3.53 | 6.65 | 6.43 |
| Median | 7.21 | 1.04 | 2.23 | 4.96 | 7.47 | 1.46 | 2.63 | 3.50 | 7.07 | 1.11 | 3.98 | 5.73 | 9.29 | 0.97 | 3.95 | 6.03 |
| Max | 13.92 | 14.71 | 13.58 | 13.04 | 30.80 | 17.59 | 15.94 | 13.67 | 14.53 | 13.70 | 18.90 | 15.46 | 17.70 | 14.14 | 18.23 | 15.00 |
| 1st Field Set | | | | | | | | | | | | | | | | |
| Mean | 12.30 | 4.64 | 3.87 | 7.91 | 9.69 | 5.17 | 4.04 | 6.12 | 10.28 | 4.01 | 3.34 | 6.46 | 11.36 | 3.85 | 3.89 | 6.65 |
| Median | 12.99 | 3.97 | 2.81 | 7.22 | 9.31 | 4.21 | 3.62 | 4.70 | 10.39 | 3.37 | 2.89 | 4.63 | 12.36 | 3.40 | 3.02 | 4.49 |
| Max | 22.07 | 16.60 | 19.78 | 22.49 | 21.61 | 18.53 | 16.48 | 18.42 | 25.64 | 14.08 | 12.69 | 18.48 | 27.93 | 14.05 | 25.11 | 18.17 |
| 2nd Field Set | | | | | | | | | | | | | | | | |
| Mean | 11.30 | 4.30 | 3.47 | 6.52 | NA | NA | NA | NA | NA | NA | NA | NA | NA | NA | NA | NA |
| Median | 9.93 | 2.81 | 2.44 | 4.86 | NA | NA | NA | NA | NA | NA | NA | NA | NA | NA | NA | NA |
| Max | 41.62 | 22.66 | 22.20 | 31.03 | NA | NA | NA | NA | NA | NA | NA | NA | NA | NA | NA | NA |

**Table 5**. Baseline Performances on 1080p Images: Mean, Median and Maximum Recovery Angular Errors of WP [3], GW [4], SoG [5] and GE [6].

| Camera Name | Canon 5DSR | | | | Nikon D180 | | | | Mobile (With CSC) | | | | Mobile (No CSC) | | | |
|---|---|---|---|---|---|---|---|---|---|---|---|---|---|---|---|---|
| Algorithms | WP | GW | SoG | GE | WP | GW | SoG | GE | WP | GW | SoG | GE | WP | GW | SoG | GE |
| Lab Printouts | | | | | | | | | | | | | | | | |
| Mean | 2.63 | 5.09 | 3.80 | 3.07 | 2.58 | 5.87 | 4.28 | 3.35 | 2.44 | 4.54 | 3.52 | 3.70 | 2.29 | 4.55 | 3.50 | 3.61 |
| Median | 2.00 | 4.15 | 2.97 | 2.38 | 2.23 | 4.82 | 3.25 | 2.65 | 1.92 | 3.58 | 2.86 | 2.71 | 1.78 | 3.61 | 2.74 | 2.68 |
| Max | 19.33 | 16.99 | 16.34 | 14.56 | 13.82 | 18.84 | 18.51 | 15.51 | 11.41 | 15.07 | 15.07 | 18.75 | 10.16 | 15.02 | 14.73 | 17.61 |
| Lab Real Scenes | | | | | | | | | | | | | | | | |
| Mean | 7.11 | 3.96 | 4.90 | 4.61 | 6.32 | 4.91 | 5.39 | 4.53 | 8.24 | 3.45 | 6.51 | 6.11 | 8.70 | 3.54 | 6.60 | 6.35 |
| Median | 8.39 | 1.04 | 3.40 | 3.68 | 7.23 | 1.45 | 4.23 | 2.46 | 9.13 | 1.11 | 3.93 | 5.46 | 10.17 | 0.97 | 3.99 | 5.76 |
| Max | 14.03 | 14.70 | 13.70 | 13.14 | 11.11 | 17.55 | 16.32 | 15.41 | 16.84 | 13.75 | 18.14 | 15.56 | 16.08 | 14.16 | 17.86 | 15.19 |
| 1st Field Set | | | | | | | | | | | | | | | | |
| Mean | 12.65 | 4.52 | 6.02 | 8.07 | 8.84 | 5.10 | 5.41 | 5.98 | 10.65 | 3.96 | 4.19 | 6.94 | 10.47 | 3.82 | 4.39 | 6.90 |
| Median | 13.98 | 3.94 | 4.89 | 5.49 | 8.95 | 4.21 | 4.40 | 3.79 | 12.21 | 3.28 | 3.31 | 4.78 | 11.61 | 3.35 | 3.37 | 4.77 |
| Max | 22.72 | 16.60 | 18.98 | 23.29 | 19.49 | 18.53 | 22.06 | 18.61 | 19.93 | 14.08 | 15.79 | 19.72 | 21.02 | 14.05 | 19.68 | 19.35 |
| 2nd Field Set | | | | | | | | | | | | | | | | |
| Mean | 10.71 | 4.21 | 4.91 | 6.98 | NA | NA | NA | NA | NA | NA | NA | NA | NA | NA | NA | NA |
| Median | 11.06 | 2.73 | 3.54 | 4.74 | NA | NA | NA | NA | NA | NA | NA | NA | NA | NA | NA | NA |
| Max | 29.00 | 22.66 | 19.29 | 24.10 | NA | NA | NA | NA | NA | NA | NA | NA | NA | NA | NA | NA |

## IV. Experimental Results

### A. Evaluation of Unsupervised Baseline Methods

In this section, we first evaluate some baseline unsupervised methods: White Patch (WP) [3], Gray World (GW) [4], Shades of Gray [5] (SoG), and Gray Edge (GE) [6]. The different algorithms' parameters were selected according to the suggestions given by the authors in the respective references. The Minkowski norm of 6 was selected for SoG and GE. The local scale parameter $\sigma$ was set to 2 and the first order differentiation was used for GE.

Given, the estimated light source chromaticity $\rho^{Est}$ and the ground truth (white point) $\rho^{E}$, the evaluation is based on the widely used recovery angular error ($RAE$) obtained as follows.

$$RAE = \cos^{-1}\left(\frac{\rho^{E}\rho^{Est}}{\|\rho^{E}\|\|\rho^{Est}\|}\right) \qquad (3)$$

Table 4 and Table 5 show the mean, median and maximum $RAE$ values of the baseline methods WP [3], GW [4], SoG [5] and GE [6], measured on the original high resolution images and the reduced 1080p images of the proposed Intel-TUT datase. A few observations that one can make from these tables are provided below.

In real lab scenes, GW usually performs favorably. Especially in 1080p images, GW often ranks on top in this set as well as in the field set. This can be explained by the fact that real lab scenes contain either a dominant gray background or a wide variety of colors in the scene, which are both in line with GW assumption that the average chromaticity in an image is gray. It should be noted here that the performance of real lab scenes might not be very representative due to the limited number of examples (20 images only); however, they are still kept as they form a unique set with real contents under controlled illumination.

WP performs well in lab printouts. Especially in 1080p images, WP usually ranks on top in this set of images. In all other datasets, WP performs the worst. The good performance in lab printouts might be due to the saturated image regions. In real images, saturated pixels can be useless; however, in printouts these points correspond to white objects from which the white point can be accurately recovered. Note that by saturated points in printouts, we mean the points that are saturated in the original capture.

SoG performs favorably in high resolution field images. Due to the specific choice of the Minkowski norm, SoG weights brighter regions higher. In case small local bright details are more often achromatic, or if there are minor non-clipped specular reflections, this would result in favorable performance of SoG. Keeping in mind that these details are more evident in high-resolution images and the fact that field images contain more content variety, resulting in additional specularities, this would explain the better performance of SoG in high resolution field images.

WP usually benefits from the resolution loss, whereas GE and SoG suffer from it. GW is the most robust algorithm to resolution change. Since WP is an algorithm that is easily effected by noise, it is reasonable that it benefits from the averaged pixel values in 1080p images. On the other hand, due to the loss of edge information in 1080p images, it is expected that GE would perform poorly in these images compared to high-resolution images. The reason why SoG would fail in low-resolution images is explained in the previous paragraph. Due to its averaging strategy, GW algorithm's robustness is straightforward. In fact, if downsampling is applied using the bilinear technique, ideally GW's performance should be identical in high and low-resolution images. However, in some images due to averaging affects, some saturated points are lost in the low-resolution images. While measuring performance, these saturated regions are discarded. Hence, this results in different performance of GW in high and low resolution images.

Algorithm performances do not show much variation in CS corrected Mobile images compared to uncorrected ones. The ground truths for the non-CSC images were set as averages of achromatic patches in different parts of the image since different locations yield different ground truths. The non-WP methods used here average their calculation results, so it can arrive at similar result on the average.

Table 4 and Table 5 evaluate individual algorithm performances and invariance against the images sets. Next, we evaluate the camera invariance of the baseline methods. To achieve this, first, the standard deviation of the $RAE$ measures of the baseline methods are calculated for each individual image from a given set over different cameras. Then, the mean, median and maximum of these standard deviations over the images in the set is calculated. Table 6 and Table 7 show this average standard deviation values for each image and for each algorithm, for the high resolution and the 1080p images, respectively. Note that the 2nd field set was not included when calculating the statistics since it does not contain images taken by all the cameras. Moreover, mobile camera without CS correction was not included in the experiments in line with our comments in Section III.E.4).

Table 6 and Table 7 illustrate that the baseline methods' performances are clearly effected by the choice of camera. The high variance of WP algorithm is mostly due to the method's unreliable nature as it relies on the brightest pixels which might be noisy. Being entirely unsupervised and somewhat a more general method compared to others (assuming the world is gray), the GW algorithm's robustness to camera change has been observed to be better than that of the other baselines. SoG and GE require parameter fine-tuning, hence these methods are not so robust to camera changes. Nevertheless, these algorithms are still relatively more general methods compared to entirely supervised methods. Thus, it is expected that if an entirely supervised method is directly applied without somehow incorporating inter-camera relations during training or testing, the performance would be more effected by the camera changes. Next, we experimentally show that this claim is valid for a supervised method that we have devised.

**Table 6.** Mean, median and maximum standard deviations ($\bar{\sigma}, \tilde{\sigma}, \sigma_M$) of baselines WP [3], GW [4], SoG [5] and GE [6] across the cameras in high resolution images.

| | WP | | | GW | | | SoG | | | GE | | |
|---|---|---|---|---|---|---|---|---|---|---|---|---|
| | $\bar{\sigma}$ | $\tilde{\sigma}$ | $\sigma_M$ | $\bar{\sigma}$ | $\tilde{\sigma}$ | $\sigma_M$ | $\bar{\sigma}$ | $\tilde{\sigma}$ | $\sigma_M$ | $\bar{\sigma}$ | $\tilde{\sigma}$ | $\sigma_M$ |
| LP | 1.27 | 0.93 | 10.69 | 0.74 | 0.62 | 2.29 | 0.68 | 0.53 | 2.48 | 1.08 | 0.62 | 9.11 |
| LR | 3.75 | 2.27 | 14.38 | 0.88 | 0.25 | 4.17 | 1.27 | 0.69 | 5.74 | 2.36 | 2.24 | 6.69 |
| F | 3.77 | 3.17 | 11.51 | 0.81 | 0.77 | 2.23 | 1.48 | 0.88 | 6.53 | 1.97 | 1.85 | 4.95 |

**Table 7.** Mean, median and maximum standard deviations of baselines WP [3], GW [4], SoG [5] and GE [6] across the cameras in 1080p images.

| | WP | | | GW | | | SoG | | | GE | | |
|---|---|---|---|---|---|---|---|---|---|---|---|---|
| | $\bar{\sigma}$ | $\tilde{\sigma}$ | $\sigma_M$ | $\bar{\sigma}$ | $\tilde{\sigma}$ | $\sigma_M$ | $\bar{\sigma}$ | $\tilde{\sigma}$ | $\sigma_M$ | $\bar{\sigma}$ | $\tilde{\sigma}$ | $\sigma_M$ |
| LP | 0.81 | 0.54 | 7.45 | 0.73 | 0.61 | 2.29 | 0.67 | 0.50 | 3.10 | 0.95 | 0.52 | 8.71 |
| LR | 2.65 | 2.28 | 7.45 | 0.86 | 0.25 | 4.10 | 1.42 | 1.17 | 5.79 | 1.31 | 1.04 | 4.76 |
| F | 2.83 | 2.60 | 8.39 | 0.79 | 0.75 | 2.36 | 1.63 | 0.96 | 7.55 | 1.79 | 1.36 | 7.24 |

**Table 8. (Camera Invariance)** Mean, median and maximum recovery angle errors ($\overline{RAE}$, $\widetilde{RAE}$, $RAE_M$ ) of training, validation and test errors for each fold –High Resolution Images: Canon (C), Nikon (N), Mobile (M). Epochs of selected models are indicated next to folds.

| | Fold 1 (98th) | | | Fold 2 (96th) | | | Fold 3 (20th) | | | Fold 4 (18th) | | | Fold 5 (8th) | | | Fold 6 (14th) | | |
|---|---|---|---|---|---|---|---|---|---|---|---|---|---|---|---|---|---|---|
| | N | M | C | M | N | C | N | C | M | M | C | N | C | N | M | C | M | N |
| $\overline{RAE}$ | 3.902 | 3.426 | 5.246 | 3.08 | 4.399 | 5.353 | 4.811 | 4.683 | 3.978 | 4.206 | 4.528 | 5.357 | 4.586 | 5.188 | 4.56 | 5.215 | 4.725 | 5.064 |
| $\widetilde{RAE}$ | 3.151 | 2.778 | 4.803 | 2.477 | 3.488 | 4.836 | 3.737 | 4.135 | 3.61 | 3.541 | 3.729 | 4.097 | 3.799 | 4.587 | 4.268 | 4.602 | 4.044 | 4.552 |
| $RAE_M$ | 17.888 | 15.68 | 16.035 | 14.178 | 16.451 | 14.782 | 20.924 | 15.549 | 18.282 | 16.248 | 14.589 | 17.022 | 15.348 | 19.344 | 16.128 | 16.34 | 14.771 | 17.867 |

**Table 9. (Camera Invariance)** Mean, median and maximum recovery angle errors ($\overline{RAE}$, $\widetilde{RAE}$, $RAE_M$ ) of training, validation and test errors for each fold -1080p Images: Canon (C), Nikon (N), Mobile (M). Epochs of selected models are indicated next to folds.

| | Fold 1 (97th) | | | Fold 2 (94th) | | | Fold 3 (63rd) | | | Fold 4 (62nd) | | | Fold 5 (49th) | | | Fold 6 (48th) | | |
|---|---|---|---|---|---|---|---|---|---|---|---|---|---|---|---|---|---|---|
| | N | M | C | M | N | C | N | C | M | M | C | N | C | N | M | C | M | N |
| $\overline{RAE}$ | 3.915 | 3.478 | 5.449 | 3.124 | 4.046 | 5.308 | 4.843 | 5.224 | 4.04 | 4.239 | 4.643 | 5.252 | 4.785 | 5.359 | 4.612 | 5.448 | 4.758 | 5.233 |
| $\widetilde{RAE}$ | 3.259 | 2.755 | 5.09 | 2.568 | 3.004 | 4.743 | 3.701 | 4.697 | 3.607 | 3.453 | 3.628 | 3.732 | 3.993 | 4.808 | 4.307 | 4.795 | 4.114 | 4.903 |
| $RAE_M$ | 16.322 | 15.863 | 14.924 | 14.412 | 16.984 | 15.159 | 20.798 | 15.948 | 18.445 | 16.256 | 14.799 | 18.331 | 15.022 | 19.095 | 16.253 | 15.972 | 14.95 | 17.612 |

## B. Evaluation of the a Baseline Supervised Method

In this section, we aim to experiment on camera invariance of a state-of-the-art Convolutional Neural Network (CNN) based direct supervised method [21], which takes local image patches and learns a regression on the white points. The CNN consists of a convolutional layer with 240 number of 1x1 filters followed rectified linear unit activation and max-pooling on 8x8 blocks with stride 8. Finally there is a fully connected layer with 40 hidden layers and outputs a 3 dimensional vector –the estimated illuminant chromaticity. We adopt the variant of CNN method in [21] where CNN is trained per patch based on the Euclidean loss. The training is conducted with 100 epochs and the model that best performs in validation set is taken. During testing, we obtain a pool of estimates for all 32x32 patches from the test images. The final estimate is simply taken by median pooling over these estimates as suggested in [21].

First, we evaluate camera invariance in the same scenes. According to suggestions in Section III.E.1), we conduct training, validation and testing in all possible six combinations. Each CNN is trained to give the least generalization error in the validation dataset.

**Table 8** illustrates the training, validation and test set errors of each fold in the high-resolution images. In each fold, the camera abbreviations (C: Canon, N: Nikon, M: Mobile) are sorted in training, validation and test set order.

One can observe that the performances on each camera differs in test set. Moreover, training, validation and test performances in each fold also are quite different. However, we observe that when the training and validation follows Nikon-Mobile or Mobile-Nikon order, there is a mutually decreasing pattern in the error. In other words, both errors keep decreasing for a large number of epochs during training. However, we do not observe this when Canon camera is either in training or validation dataset. This is evident from the epochs numbers of selected models in each fold given in **Table 8**. Given that the scenes are the same in each camera, we attribute the above observations to camera characteristics. Especially this conclusion can be validated by the similar characteristics of Nikon and Mobile and deviating characteristics of Canon cameras as observed in Fig. 7. a.

The above observations point out that the application of a direct supervised method (CNN-based in this case) that does not incorporate information about camera characteristics during training, is not camera invariant. Although one might also question the validity of the observations made in the previous paragraph due to the different camera resolutions, which lead to different number of samples (32x32 patches) available for each camera. In fact, this imbalance might affect the algorithm performance. Therefore, we have also made the above experiment on 1080p images. In this case, all images are approximately of the same resolution for all cameras. As it can be observed from Table 9, the errors are quite similar to those in **Table 8**, being very slightly higher on average. Hence, the effect of camera characteristics is evident from both experiments. Moreover, we notice that using 1080p images does not degrade the performance noticeably, at least for the CNN based approach at hand. The other experiments in this paper are only conducted in high-resolution images.

Next, we evaluate the camera and scene invariance of method in [21] according to the guidelines in Section III.E.2), i.e. we train the CNN with Nikon and Mobile images, validate on Canon images (except for the 2nd field set) and test on Canon's 2nd field images. The objective is to assess camera and scene invariance during the test. Following our observations in the previous evaluation, we do not expect the CNN model to achieve camera and scene invariances in this experiment, as it is more challenging than the previous experiment. High error rates shared in Table 10 validate our expectation.

**Table 10. (Camera and Scene Invariance)** Mean, median and maximum recovery angle errors ($\overline{RAE}$, $\widetilde{RAE}$, $RAE_M$) of Training, Validation and Test Set.

| | $\overline{RAE}$ | $\widetilde{RAE}$ | $RAE_M$ |
|---|---|---|---|
| Training | 5.426 | 4.452 | 20.324 |
| Validation | 4.160 | 3.396 | 15.116 |
| Canon 2nd Field | 6.281 | 6.408 | 24.492 |

Finally, we experiment on camera and scene invariance of the CNN-based model from single camera according to the proposed guidelines in III.E.3), i.e. we train and validate on partitions of Canon 2nd field set and test on different camera sets. This experiment aims to evaluate if the method can be trained on one camera and generalize to different scenes taken by any camera (including the one used in the training). The training is conducted in two folds using half of the images for training and half for validation. The test errors provided in Table 11 are averaged across folds. One can observe from Table 11 that training the CNN-based method on the 2nd field set cannot generalize successfully to new scenes taken by either the same camera or other cameras. However, it is evident that it performs much better in Canon images in the test set compared to images taken by other cameras. Since the training and validation is made on field images, we also evaluate field images in the test set only. We observe that, the performance on field set is much better than the rest of the dataset. This also shows the poor generalization of the method to different sets (field or lab) in the dataset.

**Table 11. (Camera and Scene Invariance from Single Camera)** Mean, median and maximum recovery angle errors ($\overline{RAE}, \widetilde{RAE}, RAE_M$) of Training, Validation and Test Set.

|  | $\overline{RAE}$ | $\widetilde{RAE}$ | $RAE_M$ |
|---|---|---|---|
| Training | 2.716 | 1.877 | 21.554 |
| Validation | 3.350 | 2.756 | 21.452 |
| Canon (Field Only) | 4.496 | 3.752 | 14.357 |
| Canon | 6.503 | 6.325 | 20.807 |
| Nikon (Field Only) | 6.122 | 5.157 | 14.911 |
| Nikon | 9.317 | 7.829 | 25.617 |
| Mobile (Field Only) | 7.188 | 5.571 | 20.841 |
| Mobile | 9.882 | 8.471 | 23.115 |

## V. Conclusion

A new Intel-TUT image database for camera invariant color constancy research is introduced in this paper. The variety of scenes and cameras, spectral power distributions of lab light sources and camera spectral responses make this dataset unique for camera invariance research. The inclusion of a mobile camera also presents a common use case as mobile cameras are more often used than DSLRs or DSCs. This dataset is also suitable for research on the effect of color shading as mobile images are provided both with corrected and uncorrected color shading.

Camera invariance is shown to be a challenging problem via observing the highly varying responses of different cameras to the same illuminating sources and the variation in the baseline algorithms' performances on different cameras. This was especially shown for a CNN-based baseline method that is evaluated in this paper. Although this model generalizes to new images in the same set (scene and camera), it is shown to be unstable when tested on new scenes and cameras. This shows the need for methods designed specifically to overcome this limitation. To evaluate performance of supervised color constancy algorithms to camera and scene invariances, guidelines are provided in this paper for a benchmark evaluation over Intel-TUT database of future color constancy methods aiming to solve the challenging camera invariance problem.

## VI. Acknowledgements

This work was supported by the NSF-TEKES CVDI sponsored by Intel Finland. We thank Vitali Samurov and Anttu Koski for providing some of the original images that were used as printouts.

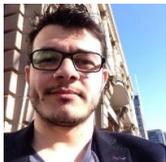

**Çağlar Aytekin** received his B.Sc. degree from the Electrical and Electronics Engineering Department from Middle East Technical University, Ankara Turkey in 2008. He received the M.Sc. department with Signal Processing major from Middle East Technical University, Ankara, Turkey in 2011. He received the Ph.D. degree from Department of Signal Processing, Tampere University of Technology in 2016. He is currently a post-doctoral researcher in Department of Signal Processing, Tampere University of Technology. He has published 19 papers in international conferences and journals. He has won the the ICPR 2014 – Best Student Paper Award for his work in unsupervised salient object detection. His research interests are, visual saliency, semantic segmentation, image processing, quantum mechanics based computer vision methods and deep learning.

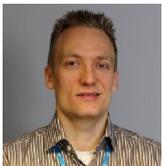

**Jarno Nikkanen** received his M.Sc. and Dr.Sc.Tech. degrees from Tampere University of Technology in 2001 and 2013, respectively, with subjects in Signal Processing and Software Systems. Jarno has 17 years of industry experience in digital imaging topics, starting at Nokia Corporation in 2000 where he developed and productized many digital camera algorithms, and moving to Intel Corporation in 2011 where he is currently working as Principal Engineer and Imaging Technology Architect. Jarno holds international patents for over 20 digital camera related inventions.

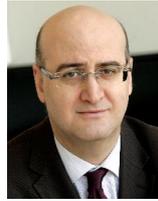

**Moncef Gabbouj** received his BS degree in electrical engineering in 1985 from Oklahoma State University, Stillwater, and his MS and PhD degrees in electrical engineering from Purdue University, West Lafayette, Indiana, in 1986 and 1989, respectively. Dr. Gabbouj is a Professor of Signal Processing at the Department of Signal Processing, Tampere University of Technology, Tampere, Finland. He was Academy of Finland Professor during 2011-2015. He held several visiting professorships at different universities. His research interests include multimedia content-based analysis, indexing and retrieval, machine learning, nonlinear signal and image processing and analysis, voice conversion, and video processing and coding. Dr. Gabbouj is a Fellow of the IEEE and member of the Academia Europaea and the Finnish Academy of Science and Letters. He is the past Chairman of the IEEE CAS TC on DSP and committee member of the IEEE Fourier Award for Signal Processing. He served as Distinguished Lecturer for the IEEE CASS. He served as associate editor and guest editor of many IEEE, and international journals. Dr. Gabbouj was the recipient of the 2015 TUT Foundation Grand Award, the 2012 Nokia Foundation Visiting Professor Award, the 2005 Nokia Foundation Recognition Award, and several Best Paper Awards.